\documentclass[runningheads]{llncs}
\usepackage[T1]{fontenc}
\usepackage{graphicx}
\usepackage{multirow}

\usepackage{amsmath}
\usepackage{bm}
\usepackage{amssymb}

\newcommand{\repeatthanks}{\textsuperscript{\thefootnote}}

\begin{document}
\title{ETSCL: An Evidence Theory-Based Supervised Contrastive Learning Framework for Multi-modal Glaucoma Grading}
\titlerunning{ETSCL: Framework for Multi-modal Glaucoma Grading}
%
\author{
Zhiyuan Yang\inst{1}\thanks{Zhiyuan Yang and Bo Zhang contributed equally.}\orcidID{0009-0007-8277-5772} \and
Bo Zhang\inst{2}\repeatthanks\orcidID{0009-0002-3736-4070} \and
Yufei Shi\inst{3}\orcidID{0000-0003-1822-1621} \and
Ningze Zhong\inst{4}\orcidID{0009-0007-6758-058X} \and
Johnathan Loh\inst{5}\orcidID{0000-0001-8176-0498} \and
Huihui Fang\inst{2}\orcidID{0000-0003-3380-7970} \and
Yanwu Xu\inst{6}\thanks{Yanwu Xu and Si Yong Yeo are the corresponding authors. Email: xuyanwu@scut.edu.cn, siyong.yeo@ntu.edu.sg}\orcidID{0000-0002-1779-931X} \and
Si Yong Yeo\inst{3}\repeatthanks\orcidID{0000-0001-6403-6019}
}

%
\authorrunning{Z. Yang et al.}

\institute{Beihang University, China \and College of Computing and Data Science, Nanyang Technological University, Singapore \and Lee Kong Chian School of Medicine, Nanyang Technological University, Singapore \and Sun Yat-sen University, China \and National University of Singapore, Singapore \and South China University of Technology, China}

\maketitle              
\begin{abstract}
Glaucoma is one of the leading causes of vision impairment. Digital imaging techniques, such as color fundus photography (CFP) and optical coherence tomography (OCT), provide quantitative and noninvasive methods for glaucoma diagnosis. Recently, in the field of computer-aided glaucoma diagnosis, multi-modality methods that integrate the CFP and OCT modalities have achieved greater diagnostic accuracy compared to single-modality methods. However, it remains challenging to extract reliable features due to the high similarity of medical images and the unbalanced multi-modal data distribution. Moreover, existing methods overlook the uncertainty estimation of different modalities, leading to unreliable predictions. To address these challenges, we propose a novel framework, namely ETSCL, which consists of a contrastive feature extraction stage and a decision-level fusion stage. Specifically, the supervised contrastive loss is employed to enhance the discriminative power in the feature extraction process, resulting in more effective features. In addition, we utilize the Frangi vesselness algorithm as a preprocessing step to incorporate vessel information to assist in the prediction. In the decision-level fusion stage, an evidence theory-based multi-modality classifier is employed to combine multi-source information with uncertainty estimation. Extensive experiments demonstrate that our method achieves state-of-the-art performance. The code is available at \url{https://github.com/master-Shix/ETSCL}.

\end{abstract}
\section{Introduction}
\label{sec:intro}
\begin{figure}[t]
  \centering
  \includegraphics[width=1\linewidth]{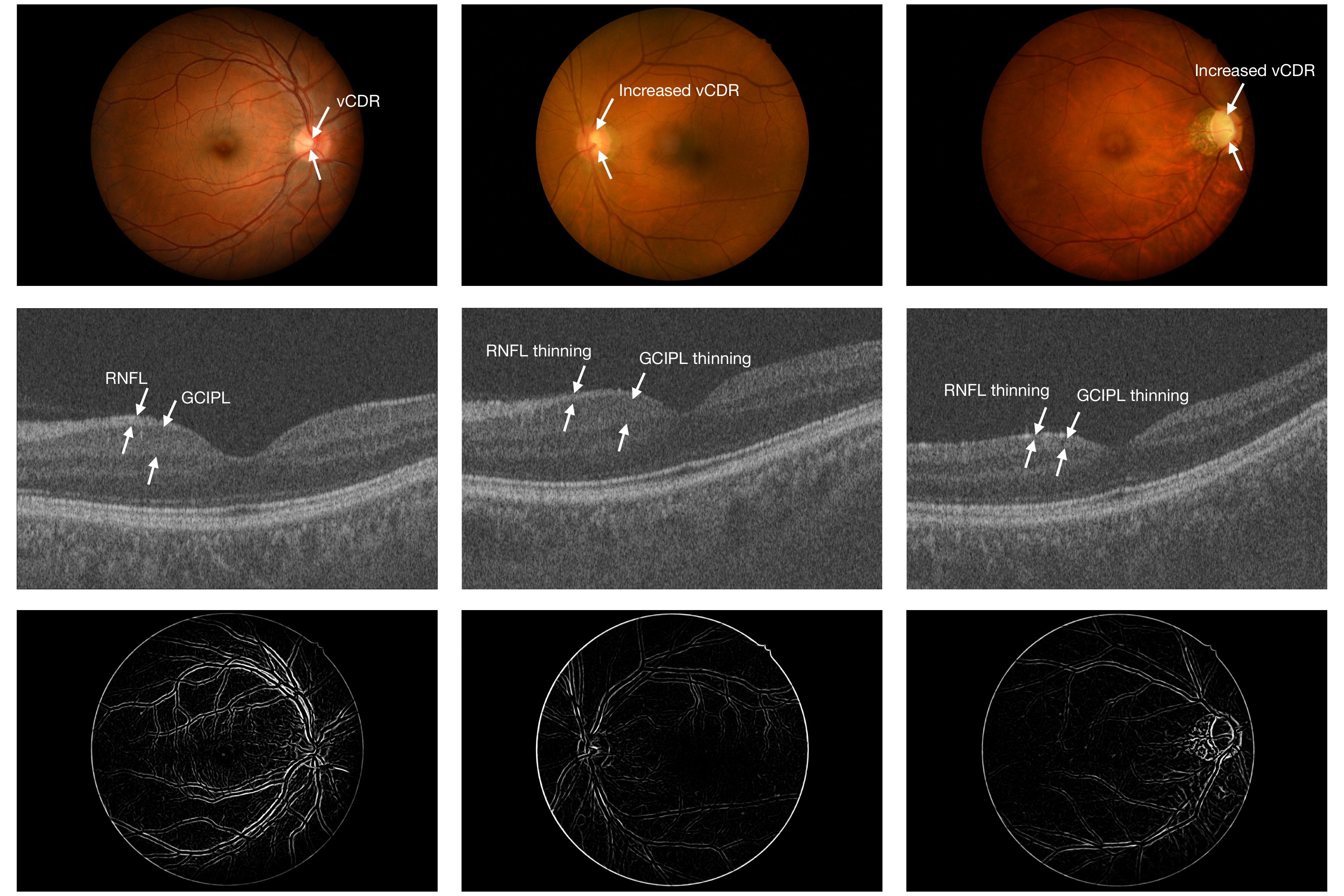}
  \begin{minipage}{0.33\linewidth}
    \centering
    \fontsize{9pt}{10pt}\selectfont
    (a) Normal
  \end{minipage}%
  \begin{minipage}{0.33\linewidth}
    \centering
    \fontsize{9pt}{10pt}\selectfont
    (b) Early
  \end{minipage}%
  \begin{minipage}{0.33\linewidth}
    \centering
    \fontsize{9pt}{10pt}\selectfont
    (c) Advanced
  \end{minipage}%
  \caption{Multi-modal imaging from the GAMMA dataset demonstrates glaucoma progression across three stages. Column (a) features images without glaucoma. Column (b) features early-stage glaucoma. Column (c) features intermediate-advanced-stage glaucoma. The top row displays CFP images, showing increased vCDR. The middle row displays OCT samples, showing thinning of the RNFL and GCIPL. The bottom row displays vessel structures, extracted from CFP images.}
  \label{fig:1}
\end{figure}

Glaucoma, a leading cause of vision impairment, causes progressive damage to the optic nerve, including the optic nerve head (ONH), the retinal nerve fibre layer (RNFL) and the ganglion cell-inner plexiform layer (GCIPL) \cite{weinreb2016primary}. Digital imaging techniques, such as color fundus photography (CFP) \cite{besenczi2016review} and optical coherence tomography (OCT) \cite{huang1991optical,de2018clinically}, offer quantitative and noninvasive ways to evaluate the optic nerve structure for glaucoma diagnosis. In Fig.~\ref{fig:1} (top and middle), we showcase detailed CFP and OCT samples, demonstrating the glaucoma progression across three stages and highlighting the evolution of key clinical measures, such as the vertical Cup-to-Disc Ratio (vCDR) and the thickness of the RNFL and GCIPL.

In the field of computer-aided glaucoma diagnosis, previous studies have primarily focused on using either CFP \cite{al2017automated,fang2022refuge2,he2022joined} or OCT \cite{asaoka2017validating,maetschke2019feature,wang2023octformer} individually. However, there is less exploration in multi-modality methods that utilize both modalities simultaneously. The GAMMA challenge \cite{wu2023gamma}, which introduces the first public multi-modality dataset, aims to advance the research in multi-modality glaucoma grading using both CFP and OCT. MM-MIL \cite{li2021MMMIL} employs over-sampling to augment CFP data, aiming to balance the multi-modal data distribution. Corolla \cite{corolla} introduces the extraction of the retinal thickness map as an alternative modality, leading to more efficient calculations and reduced memory usage. MM-RAF \cite{zhou2023miccai} develops a feature-level fusion strategy that facilitates cross-modality correlation, with an emphasis on leveraging spatial interaction between modalities through co-attention mechanism. Although existing methods \cite{wu2023gamma,li2021MMMIL,corolla,zhou2023miccai} have demonstrated impressive performance, they still face several challenges: First, current methods simply concatenate feature embeddings from separate feature encoders in the final stage, overlooking the unique uncertainty estimation inherent to each modality. The approach that treats all modalities with equal weight can result in unreliable predictions. Furthermore, given the inherently high similarity of medical images and the scarcity of labeled medical datasets, standard supervised learning methods often produce sub-optimal and inconsistent feature representations.

To address the challenges discussed above, we propose a novel framework, namely ETSCL, which consists of a contrastive feature extraction stage and a decision-level fusion stage. In the contrastive feature extraction stage, we employ the supervised contrastive loss to enhance the discriminative power, resulting in more effective feature extraction. In addition, we utilize the Frangi vesselness algorithm \cite{frangi1998multiscale} as a preprocessing step to incorporate previously neglected vessel information as a third complementary branch alongside the CFP and OCT branches. In the decision-level fusion stage, we propose a multi-modality classifier based on the evidence theory \cite{shafer1976mathematical,dempster2008upper}. The proposed classifier uses the belief functions to quantitatively estimate the uncertainty of information from different modalities, and combines the information based on Dempster's rule of combination. To the best of our knowledge, our framework represents the first attempt of applying decision-level fusion and vessel information in the multi-modality glaucoma grading problem. Extensive experiments on the public GAMMA dataset \cite{wu2023gamma} demonstrate that our method achieves state-of-the-art (SOTA) performance.

\section{Methodology}
\label{sec:method}
\begin{figure}[t]
  \centering
  \includegraphics[width=0.9\linewidth]{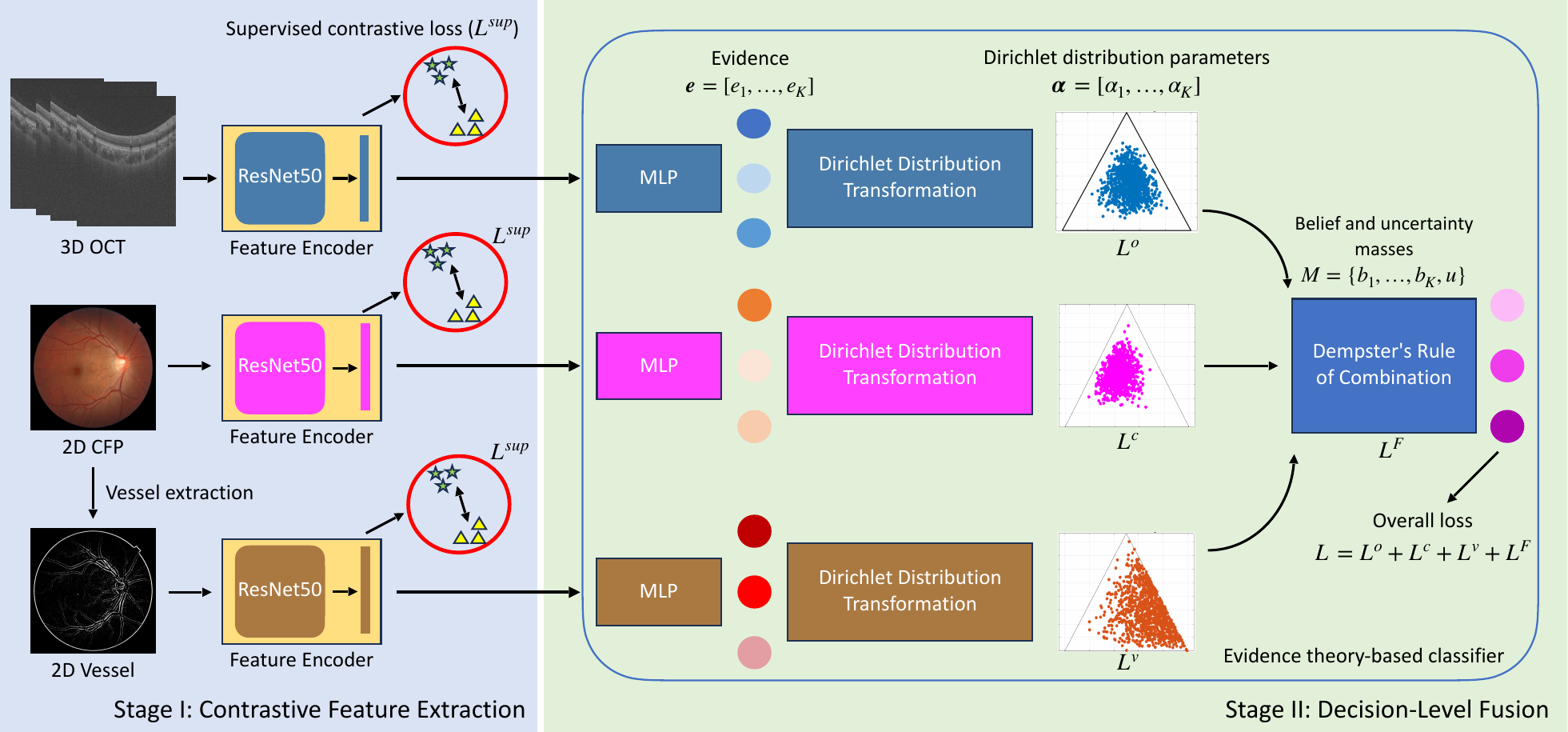}
  \caption{Overview of the proposed ETSCL framework.}
  \label{fig:2}
\end{figure}

As shown in Fig.~\ref{fig:2}, the framework is trained in two stages. In the contrastive feature extraction stage, alongside the CFP and OCT branches, vessel information is extracted from the CFP images as a third complementary branch. Vascular genesis is considered a key triggering factor in the development of glaucoma \cite{cherecheanu2013ocular,wang2015optic}. However, vessel information has been overlooked in previous glaucoma grading methods. As a preprocessing step, we employ the Frangi filter \cite{frangi1998multiscale} to extract vessel information from the CFP images. Fig.~\ref{fig:1} (bottom) illustrates the extracted vessel structures. Then the supervised contrastive loss is employed to train the three feature encoders separately, generating reliable feature embeddings. In the decision-level fusion stage, we develop an evidence theory-based classifier to fuse information from multiple sources, utilizing uncertainty estimation across different modalities.

\subsection{Contrastive Feature Extraction}
In clinical practice, extracting glaucoma-indicative features from CFP and OCT images using traditional convolutional neural networks is challenging due to the high similarity of medical images and the unbalanced multi-modal data distribution. Compared to traditional supervised learning, contrastive learning \cite{chen2020simple,henaff2020data,khosla2020supervised,chen2021empirical} leverages the semantic similarity between data points to generate more stable representations for downstream tasks. This strategy has influenced the development of existing methods such as Corolla \cite{corolla} and MM-RAF \cite{zhou2023miccai}. As inspired by the work of \cite{khosla2020supervised}, we implement a supervised contrastive learning strategy. Given $N$ randomly sampled pairs $\{\, x_h, y_h \,\}_{h=1}^{N}$, the augmented batch for contrastive learning comprises $2N$ pairs $\{\, \tilde{x}_h, \tilde{y}_h \,\}_{h=1}^{2N}$, where $y_h=\tilde{y}_{2h}=\tilde{y}_{2h-1}$. Let $i \in I = \{1, \ldots, 2N\}$ be the index of an augmented sample. The \textit{positive} is the index of the other augmented sample derived from the same sample $i$. The embedding vectors $\boldsymbol{z}$, with $\boldsymbol{z_i}$ representing the embedding vector for sample $i$, are derived via the encoder and projection head. The supervised contrastive loss \cite{khosla2020supervised} is generalized from the self-supervised contrastive loss \cite{chen2020simple,henaff2020data}, and takes the form of:
\begin{align}
L^{sup} &= \sum_{i \in I} \frac{-1}{\left|P(i)\right|} \sum_{p \in P(i)} \log \frac{\exp(\boldsymbol{z_i} \cdot \boldsymbol{z_p} / \tau)}{\sum_{a \in A(i)} \exp(\boldsymbol{z_i} \cdot \boldsymbol{z_a} / \tau)},
\label{eq:1}
\end{align}
where the $\cdot$ symbol denotes the inner dot product, $\tau$ is a scalar temperature parameter, $A(i) = I \setminus \{i\}$, and $P(i) = \{ p \in A(i) : \tilde{y}_p = \tilde{y}_i \}$ is the set of indices of all the positives in the augmented batch. The supervised contrastive loss leverages label information to generate the positives, making it more effective at grouping features of the same class closely together and separating those of different classes.

\subsection{Decision-Level Fusion}
After feature extraction, the information from different modalities is integrated in the classifier. The evidence theory, also known as the Dempster-Shafer Theory (DST) \cite{shafer1976mathematical,dempster2008upper}, introduces the concept of \textit{belief} to quantify the uncertainty of evidence. Informed by recent developments in DST-based multi-view classification \cite{han2023multiview,liu2023safe}, we design a multi-modality classifier to integrate multi-source information more effectively. In the decision-level fusion stage, the output of the multilayer perceptron (MLP) is regarded as \textit{evidence}, denoting the support for each class. For each modality, we employ the Softplus function on the feature embeddings to obtain the evidence values of all classes. The Dirichlet distribution is adopted to produce more reliable class probabilities for each modality, with the loss function designed accordingly. Subsequently, Dempster's rule of combination is employed to integrate multiple Dirichlet distributions from different modalities.

\subsubsection{Dirichlet distribution}
Given $N$ labeled samples $\{ (x_1, y_1), \ldots, (x_N, y_N) \}$, conventional classifiers employ a Softmax layer to estimate the class probabilities. Mathematically, the Softmax layer's outputs can be interpreted as parameters for a multinomial distribution. For a sample with index $i$, the probabilities of $K$ classes are $\boldsymbol{\mu_i} = [\mu_{i1}, \ldots, \mu_{iK}]$. The optimization process of Maximum Likelihood Estimation (MLE), with respect to the network's parameters $\theta$, is defined as follows:
\begin{align}
\arg\max_{\theta} \mathrm{Mult}(\mathbf{Y}; \theta) = \prod_{i=1}^{N} \prod_{j=1}^{K} \mu_{ij}^{y_{ij}},
\label{eq:2_1}
\end{align}
where $y_{ij}$ is an indicator that is 1 if the sample $i$ belongs to class $j$ and 0 otherwise. Minimizing the log-likelihood, an approach widely known as the \textit{cross-entropy loss}, essentially mirrors the objective of MLE. Given its limitations as a frequentist technique, MLE often struggles to effectively handle varying conditions and small data scenarios, which can result in overconfidence in the estimates. Alternatively, we employ the Dirichlet distribution as the conjugate prior for the multinomial distribution, allowing for more reliable and robust statistical inference. For the sample with index $i$, Dirichlet distribution is defined by $K$ parameters $\boldsymbol{\alpha_i} = [\alpha_{i1}, \ldots, \alpha_{iK}]$. The probability density function of the Dirichlet distribution is:
\begin{align}
\mathrm{Dir}(\boldsymbol{\mu_i} \mid \boldsymbol{\alpha_i}) = 
\begin{cases} 
\frac{1}{B(\boldsymbol{\alpha_i})} \prod_{j=1}^{K} \mu_{ij}^{\alpha_{ij} - 1} & \text{for } \boldsymbol{\mu_i} \in S_{unit}^{K-1}, \\
0 & \text{otherwise},
\end{cases}
\label{eq:2}
\end{align}
where $S_{unit}^{K-1}$ denotes the $K-1$ dimensional unit simplex, and $B(\boldsymbol{\alpha_i})$ represents the $K$-dimensional multivariate beta function. Consequently, label prediction for sample $i$ is modeled as a generative process:
\begin{align}
\boldsymbol{\mathbf{y}_i} \sim \mathrm{Mult}(\boldsymbol{\mathbf{y}_i} \mid \boldsymbol{\mu_i}), \quad \boldsymbol{\mu_i} \sim \mathrm{Dir}(\boldsymbol{\mu_i} \mid \boldsymbol{\alpha_i}).
\label{eq:3}
\end{align}

To design the loss function for the Dirichlet distribution of a single sample with index $i$ and label $\boldsymbol{\mathbf{y}_i}$, given the Dirichlet distribution parameters $\boldsymbol{\alpha_i}$, the Mean Square Error (MSE) loss is derived as follows:
\begin{align}
\mathrm{MSE}_i
    &= \int_{\boldsymbol{\mu_i} \in S_{unit}^{K-1}} \| \boldsymbol{\mathbf{y}_i} - \boldsymbol{\mu_i} \|_2^2 \cdot
    \mathrm{Dir}(\boldsymbol{\mu_i} \mid \boldsymbol{\alpha_i})
    d\boldsymbol{\mu_i}
    = \mathbb{E} \left[ \sum_{j=1}^{K} \left(y_{ij} - \mu_{ij}\right)^2 \right]
    \notag \\
    &= \sum_{j=1}^{K} \left( y_{ij}^2 - 2y_{ij} \mathbb{E} \left[\mu_{ij}\right] + \mathbb{E} \left[\mu_{ij}^2\right] \right)
    = \sum_{j=1}^{K} \left( y_{ij} - \mathbb{E} \left[\mu_{ij}\right] \right)^2 +  \mathrm{Var}(\mu_{ij})
    \notag \\
    &= \sum_{j=1}^{K} \left( y_{ij} - \frac{\alpha_{ij}}{S_i} \right)^2 + \frac{\alpha_{ij} \left(S_i - \alpha_{ij}\right)}{S_i^2 \left(S_i + 1\right)},
\label{eq:4}
\end{align}
where $S_i = \sum_{j=1}^{K} \alpha_{ij}$, known as the Dirichlet strength. To counteract any drift towards an unrealistic distribution, a penalty term derived from the Kullback-Leibler (KL) divergence, is incorporated to adjust the Dirichlet distribution of the sample $i$ accordingly. Consequently, the loss function for this Dirichlet distribution is:
\begin{align}
L_i = \mathrm{MSE}_i + \lambda_i \cdot \mathrm{KL}[\mathrm{Dir}(\boldsymbol{\mu_i} \mid \boldsymbol{\alpha_i}) \,||\, \mathrm{Dir}(\boldsymbol{\mu_i} \mid \mathbf{1})],
\label{eq:5}
\end{align}
where $\mathbf{1}$ represents a $K$-dimensional unit vector, and $\lambda_i$ is the annealing coefficient.

\subsubsection{Belief, uncertainty and combination}
Subjective logic (SL) \cite{jsang2018subjective} provides a framework that links the Dirichlet distribution parameters to the belief and uncertainty associated with each class. For the Dirichlet distribution of a single sample, SL assigns a belief mass $b_j$ to class $j$, reflecting the confidence in the occurrence of each class. An aggregate uncertainty mass $u$ quantifies the overall uncertainty across all classes. The summation of belief masses of all classes and the uncertainty mass equals 1, ensuring a normalized representation of belief and uncertainty. Given the evidence $e_j$ for class $j$, the Dirichlet distribution parameter, $\alpha_j$, is calculated as $\alpha_j=e_j+1$. The belief masses $\mathbf{b}$ and the uncertainty mass $u$ are derived from the Dirichlet distribution parameters as follows:
\begin{align}
\mathbf{b} = \frac{\mathbf{e}}{S} = \frac{\boldsymbol{\alpha} - 1}{S}, \quad u = \frac{K}{S},
\label{eq:6}
\end{align}
where $\boldsymbol{\alpha} - 1$ indicates the subtraction of 1 from each parameter in the Dirichlet distribution parameter vector $\boldsymbol{\alpha}$, and $S = \sum_{j=1}^{K} (e_j + 1) = \sum_{j=1}^{K} \alpha_j$, known as the Dirichlet strength. We then adopt a reduced Dempster's combination rule \cite{han2023multiview} to manage computational complexity by avoiding the large numbers of input and output masses in the original rule. For the fusion of the CFP and OCT branches, the mass set of the fusion $M^F = \{b_1^F, \ldots, b_K^F, u^F\}$ is calculated from the mass sets of the CFP branch $M^c = \{b_1^c, \ldots, b_K^c, u^c\}$ and the OCT branch $M^o = \{b_1^o, \ldots, b_K^o, u^o\}$ as follows:
\begin{align}
b_j^F = \frac{1}{1 - K'}(b_j^c b_j^o + b_j^c u^o + b_j^o u^c), u^F=\frac{1}{1 - K'} u^c u^o,
\label{eq:7}
\end{align}
where $K'=\sum_{j \neq i}b_{j}^c b_{i}^o$ is the conflict measure between the CFP and OCT branches. Then we apply the same computation to combine this CFP and OCT fusion $M^F$ with the Vessel branch to calculate the final mass set. Subsequently, based on the final mass set, we can obtain the corresponding parameters of the joint Dirichlet distribution.

\subsubsection{Evidence theory-based classifier}
After deriving the joint Dirichlet distribution from different modalities, the class with the highest probability is identified as the final prediction output. Based on Eq. (\ref{eq:5}), the loss for each modality is calculated as the summation of losses for all samples: $\sum_{i=1}^{N} L_i$. Consequently, we have the losses for the CFP, OCT, and Vessel branches, denoted as $L^c$, $L^o$, and $L^v$, respectively. Additionally, we sum up the loss of the joint Dirichlet distribution across all samples to calculate the fusion loss, $L^F$. The overall loss function of the classifier is then calculated as the summation of the losses from all the modalities and the fusion loss, as illustrated in Fig.~\ref{fig:2}.

\section{Experiments}
\label{sec:experiment}

\subsection{Dataset}
The public GAMMA dataset \cite{wu2023gamma} consists of 300 pairs of CFP images and OCT volumes, with each OCT volume consisting of 256 2D slices. The dataset is categorized into three labels: non-glaucoma, early-glaucoma, and intermediate-advanced-glaucoma, presenting a multiclass classification task. We divide the dataset into 200 samples for training and 100 samples for testing.

\subsection{Implementation Details}
All experiments are conducted on two NVIDIA A40 GPUs. For image augmentation, we apply normalization, color jittering, random grayscale conversion, center cropping, and random horizontal flipping. In the feature extraction stage, we utilize ResNet50, pretrained on ImageNet, as the backbone for all the three branches. We process the 3D OCT inputs as 256-channel 2D images. For each feature encoder, we then employ 2 fully-connected layers with ReLU activation to generate a 128-dimensional tensor, which is subsequently used for the supervised contrastive loss $L^{sup}$, with the temperature parameter $\tau$ set to 0.05. The network is trained using Adam optimizer, with a learning rate of 0.001 and a batch size of 14, over 10 epochs. In the decision-level fusion stage, the embeddings generated by each feature encoder from the previous stage are independently passed to separate MLP networks. Each MLP, with an input size of 128 and an output size of 3, is trained using the Adam optimizer over 200 epochs.

\subsection{Results and Ablation Study}
Following the evaluation approach in the GAMMA challenge, we use Cohen's kappa coefficient as the primary metric and accuracy as the secondary metric. For this ordinal ternary classification task, kappa is quadratically weighted to reflect the varying degrees of disagreement across ordinal categories \cite{wu2023gamma}, which is particularly effective in assessing models on unbalanced datasets.

We select both single-modality and multi-modality methods as the baselines. For the single-modality baselines, ResNet50 is selected for CFP, while ResNet50 and 3D EfficientNet are selected for OCT. For the multi-modality baselines, we choose a dual-branch ResNet50, and a hybrid of ResNet50 for CFP and 3D EfficientNet for OCT. Additionally, the GAMMA challenge winner's method is selected as another multi-modality baseline. As shown in Table~\ref{tab:table1}, our method outperforms the best baseline, the GAMMA challenge winner, by over 3\% in kappa and by 3\% in accuracy.
\begin{table*}[tp]
\centering
\caption{Performance of the baselines and our ETSCL on the GAMMA dataset.}
\begin{tabular}{l|l|cc}
\hline
\multicolumn{2}{c|}{Method} & Kappa & Accuracy \\
\hline
\multirow{3}{*}{Single-modality} & CFP - ResNet & 0.6199 & 0.68 \\
                                 & OCT - ResNet & 0.5006 & 0.62 \\
                                 & OCT - 3D EfficientNet & 0.4930 & 0.60 \\
\hline
\multirow{4}{*}{Multi-modality} & CFP+OCT - Dual ResNet & 0.7129 & 0.75 \\
                                & CFP+OCT - ResNet+3D EfficientNet & 0.7200 & 0.75 \\
                                & GAMMA challenge winner \cite{wu2023gamma} & 0.8535 & 0.81 \\
                                & ETSCL & \textbf{0.8844} & \textbf{0.84} \\
\hline
\end{tabular}
\label{tab:table1}
\end{table*}

\begin{table*}[tp]
\centering
\caption{Ablation study results on the GAMMA dataset. CFP means the presence of the CFP branch. OCT means the presence of the OCT branch. Vessel denotes the addition of the Vessel branch. SCL indicates the use of the supervised contrastive loss. ET represents the evidence theory-based classifier.}
\begin{tabular}{ccccc|cc}
\hline
CFP & OCT & Vessel & SCL & ET & Kappa & Accuracy \\
\hline
\checkmark & & & & & 0.6199 & 0.68 \\
\checkmark & \checkmark & & & & 0.7129 & 0.75 \\
\checkmark & \checkmark & \checkmark & & & 0.7478 & 0.77 \\
\checkmark & & & \checkmark & & 0.8170 & 0.77 \\
\checkmark & \checkmark & & \checkmark & & 0.8273 & 0.80 \\
\checkmark & \checkmark & \checkmark & \checkmark & & 0.8592 & 0.81 \\
\checkmark & \checkmark & \checkmark & \checkmark & \checkmark & \textbf{0.8844} & \textbf{0.84} \\
\hline
\end{tabular}
\label{tab:table2}
\end{table*}

As shown in Table~\ref{tab:table2}, the improvements observed when adding the supervised contrastive loss to conventional CFP, dual branches (CFP and OCT), and triple branches (CFP, OCT, and Vessel) demonstrate the significant contribution of the supervised contrastive loss. Furthermore, comparisons between dual branches (CFP and OCT) and triple branches (CFP, OCT, and Vessel), both with and without the supervised contrastive loss, validate the benefit of incorporating the vessel information. Lastly, the advancement from using the conventional Softmax layer to using the evidence theory-based classifier in the triple branches (CFP, OCT, and Vessel) highlights the superior performance of the evidence theory-based classifier.

In Table~\ref{tab:table3}, we present a comparison of our ETSCL with other SOTA methods that were published after the GAMMA challenge. We include only the kappa values as reported in their respective publications, since not all studies disclose their accuracy. Among these published SOTA methods, our ETSCL achieves the highest kappa. It is worth noting that this comparison might not be strict, due to variations in the datasets used by each method.

\begin{table*}[tp]
\centering
\caption{Comparison between our ETSCL and other SOTA methods on the GAMMA dataset. $\dagger{}$ Corolla uses 100 samples for training and testing. * MM-RAF trains on a private dataset and tests with 100 GAMMA samples on both its method and MM-MIL.}
\begin{tabular}{l|c}
\hline
Method & Kappa \\
\hline
Corolla\textsuperscript{$\dagger{}$} \cite{corolla} & 0.8550 \\
MM-MIL\textsuperscript{*} \cite{li2021MMMIL} & 0.8562 \\
MM-RAF\textsuperscript{*} \cite{zhou2023miccai} & 0.8467 \\
ETSCL & \textbf{0.8844} \\
\hline
\end{tabular}
\label{tab:table3}
\end{table*}

\section{Conclusion}
\label{sec:conclusion}
In this study, we propose an evidence theory-based supervised contrastive learning framework for multi-modality glaucoma grading. The supervised contrastive loss is employed to extract effective feature embeddings. To integrate multi-modality information with uncertainty estimation, we design an evidence theory-based classifier. In addition, vessel information is utilized to assist in the prediction. However, the limited size of the GAMMA dataset constrains our method choices, such as Vision Transformer which often leads to overfitting. The lack of private datasets restricts our ability to further test the generalizability and robustness of our method, as the GAMMA dataset is currently the only high-quality public dataset with multi-stage glaucoma grading. Furthermore, subsequent generalizability tests can reinforce the validation of the Vessel branch’s efficacy, considering the novelty of incorporating vessel information.

\begin{credits}
\subsubsection{\ackname} This project is supported by the Ministry of Education, Singapore, under its Academic Research Fund Tier 1 (RS16/23).

\subsubsection{\discintname}
The authors have no competing interests to declare that are relevant to the content of this article.
\end{credits}

%
%
%

\bibliographystyle{splncs04_modified}
\bibliography{my_reference}

\end{document}